\RequirePackage{amsmath}
\documentclass[runningheads]{llncs}
\usepackage[T1]{fontenc}
\usepackage{amsfonts}
\usepackage{graphicx}
\usepackage{fontawesome}
\usepackage{booktabs}
\usepackage{censor}
\DeclareMathOperator*{\argmax}{argmax}
\newcommand{\figref}[1]{\figurename~\ref{#1}}
\newcommand{\tabref}[1]{\tablename~\ref{#1}}
\newcommand{\secref}[1]{Section~\ref{#1}}

\usepackage{hyperref}
\usepackage{color}

\usepackage{pdfpages} 
\usepackage{pgffor} 

\makeatletter
\AtBeginDocument{\let\LS@rot\@undefined}
\makeatother

\def\supplementfilename{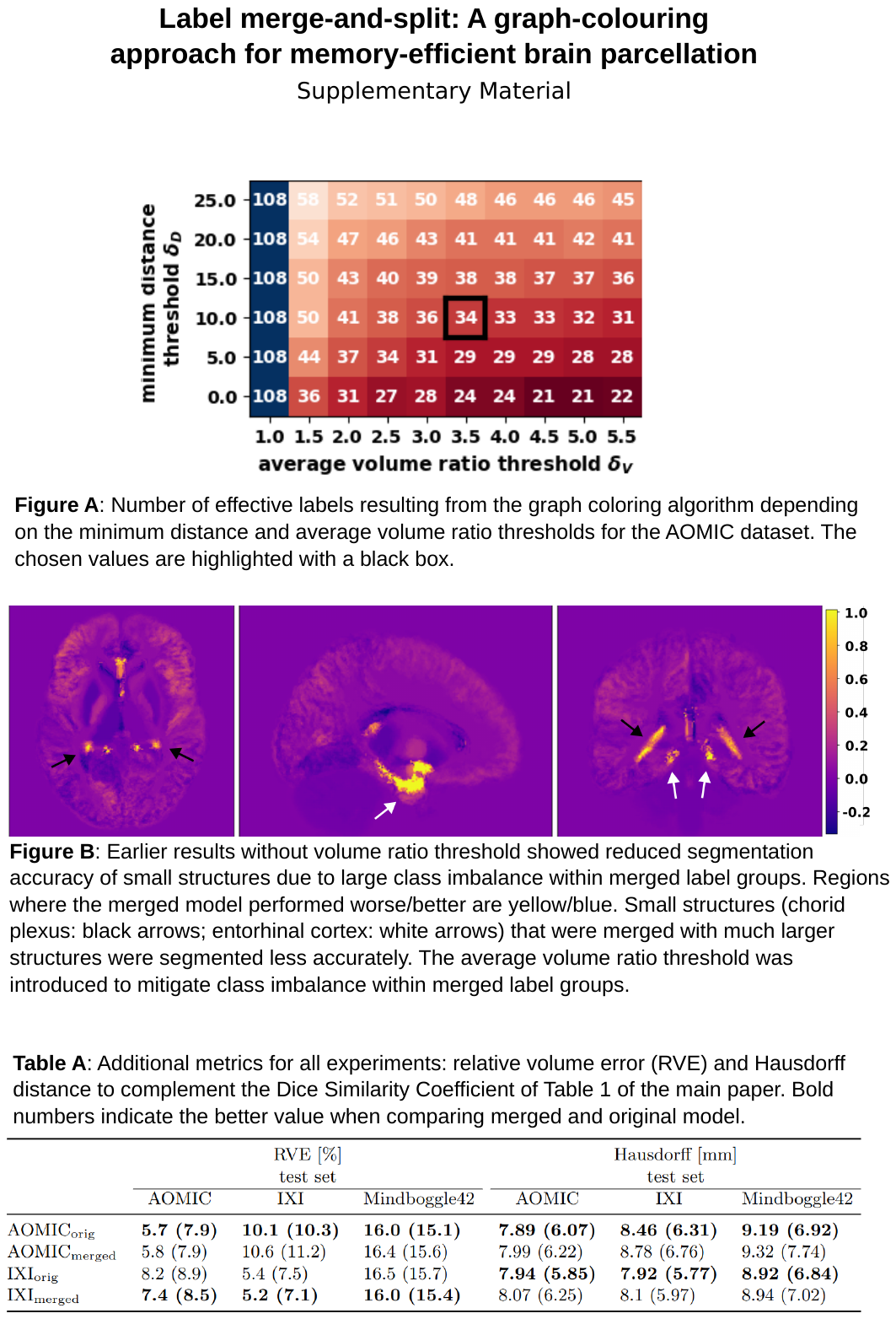}

\pdfximage{\supplementfilename}
\def\numbersupplementpages{\the\pdflastximagepages}

\newif\ifarXiv
\arXivtrue 
\begin{document}
\title{Label merge-and-split: A graph-colouring approach for memory-efficient brain parcellation}
\titlerunning{Label merge-and-split}
%
\author{Aaron Kujawa\inst{1}\orcidID{0000-0002-1445-6207} \and Reuben Dorent\inst{2}\orcidID{0000-0002-7530-0644} \and Sebastien Ourselin\inst{1}\orcidID{0000-0002-5694-5340} \and Tom Vercauteren\inst{1}\orcidID{0000-0003-1794-0456}}
\authorrunning{A. Kujawa et al.}
%
\institute{ King's College London, UK \and Harvard University, USA}
\maketitle              
\begin{abstract}
Whole brain parcellation requires inferring hundreds of segmentation labels in large image volumes and thus presents significant practical challenges for deep learning approaches.
We introduce label merge-and-split, a method that first greatly reduces the effective number of labels required for learning-based whole brain parcellation and then recovers original labels.
Using a greedy graph colouring algorithm, our method automatically groups and merges multiple spatially separate labels prior to model training and inference. The merged labels may be semantically unrelated. 
A deep learning model is trained to predict merged labels. At inference time, original labels are restored using atlas-based influence regions. 
In our experiments, the proposed approach reduces the number of labels by up to 68\% while achieving segmentation accuracy comparable to the baseline method without label merging and splitting. 
Moreover, model training and inference times as well as GPU memory requirements were reduced significantly.
The proposed method can be applied to all semantic segmentation tasks with a large number of spatially separate classes within an atlas-based prior.

\end{abstract}

\section{Introduction}
Whole brain parcellation (WBP) is a multi-class semantic segmentation task that requires the assignment of labels to all brain regions based on distinct anatomical and/or functional features.
Although the level of granularity may vary, WBP typically requires dealing with hundreds of distinct classes.
For example, the parcellation tools Freesurfer \cite{fischl2002whole} and GIF (Geodesic Information Flows) \cite{cardoso2015geodesic,prados2016niftyweb} predict up to 105 classes and 158 classes, respectively.
While classical tools such as Freesurfer and GIF are robust and accurate, processing times vary between multiple hours to a full day for a single input image. 

Deep learning models can greatly reduce inference time, with some methods achieving WBP within 1 minute \cite{henschel2020fastsurfer,roy2017error,roy2019quicknat} or as little as 11 seconds \cite{roy20222}.
An important limitation of existing approaches
is the high demand for GPU and CPU memory and computational power. Bottleneck operations are typically those which involve continuous values/probabilities for each label such as multi-label resampling (for image scaling and spatial data augmentation) and label-wise loss calculations. Resources required for these operations scale linearly with the number of labels.

While smaller network architectures and techniques like network compression and model quantization have been employed to partially address this problem \cite{li2017compactness},
the fixed output size of the softmax layer at the end of 3D networks 
remains a driver of GPU requirements and thus a bottleneck for all deep neural networks.

This has led to mitigation strategies such as
patch-based methods which extract smaller sub-volumes from the full image during training and inference, at the cost of losing spatial context \cite{moeskops2016automatic,mehta2017brainsegnet,li2017compactness,dolz20183d,wachinger2018deepnat,huo20193d,coupe2019assemblynet,roy20222,mehta2017brainsegnet}.
Alternatively, 2.5D methods reduce memory requirements by producing separate predictions for all 2D slices in axial, coronal and sagittal orientation \cite{henschel2020fastsurfer,roy2017error,roy2019quicknat}. 

Other existing approaches further reduce memory and computational requirements by merging semantically related label pairs during model training and inference \cite{henschel2020fastsurfer,roy20222}.
These methods exploit the symmetry of pairs of structures that are present in both brain hemispheres, for example the right and left Pallidum, and merge the corresponding labels into a single label.
At inference time, as a post-processing step, the predicted merged label is separated along the mid-sagittal plane to recover the original labels.
However, many brain structures do not have left/right pairs (e.g., brain stem).
In other cases, the left/right components may be spatially too close to the mid-sagittal plane (e.g., left/right cerebellum white matter), which makes their separation during post-processing error-prone.
This limits the reduction in the number of labels to a relatively small fraction. For example, in \cite{henschel2020fastsurfer} and \cite{roy20222}, a reduction from 95 to 78 labels was achieved by merging 17 label pairs. The process also requires manual specification of symmetric and separate label pairs.

In this work, we propose an automated and scalable method to significantly reduce the number of labels while allowing for accurate label splitting during post-processing. The method includes the following steps: Firstly, we construct an undirected graph where vertices correspond to brain structures and edges to discretised spatial distance priors. Using a graph colouring algorithm \cite{matula1983smallest}, our approach automatically groups and merges multiple spatially separate labels of similar volume. The spatial separation is required to allow for label splitting during inference, while the constraint for similar volumes prevents high class imbalance. Secondly, a memory-efficient 3D U-Net approach is trained for automated parcellation and prediction of merged labels. Thirdly, we employ a strategy to split the predicted merged segmentation labels using atlas-based influence regions. Finally, experiments on three publicly available datasets (Mindboggle101 \cite{klein2012101}, AOMIC \cite{snoek2021amsterdam}, and IXI \cite{ixi} ) demonstrate the effectiveness of the proposed merge-and-split approach. The number of labels is reduced by up to 68\%, resulting in a great reduction of training and inference time (by 43-49\%) and GPU requirements during training (by 50\%) and inference (by 21\%) while achieving a segmentation quality comparable to the memory-heavy model trained without label merging and splitting.


\section{Methods}
\label{sec:methods}
\begin{figure}[htb]
\centering
\includegraphics[width=0.8\textwidth]{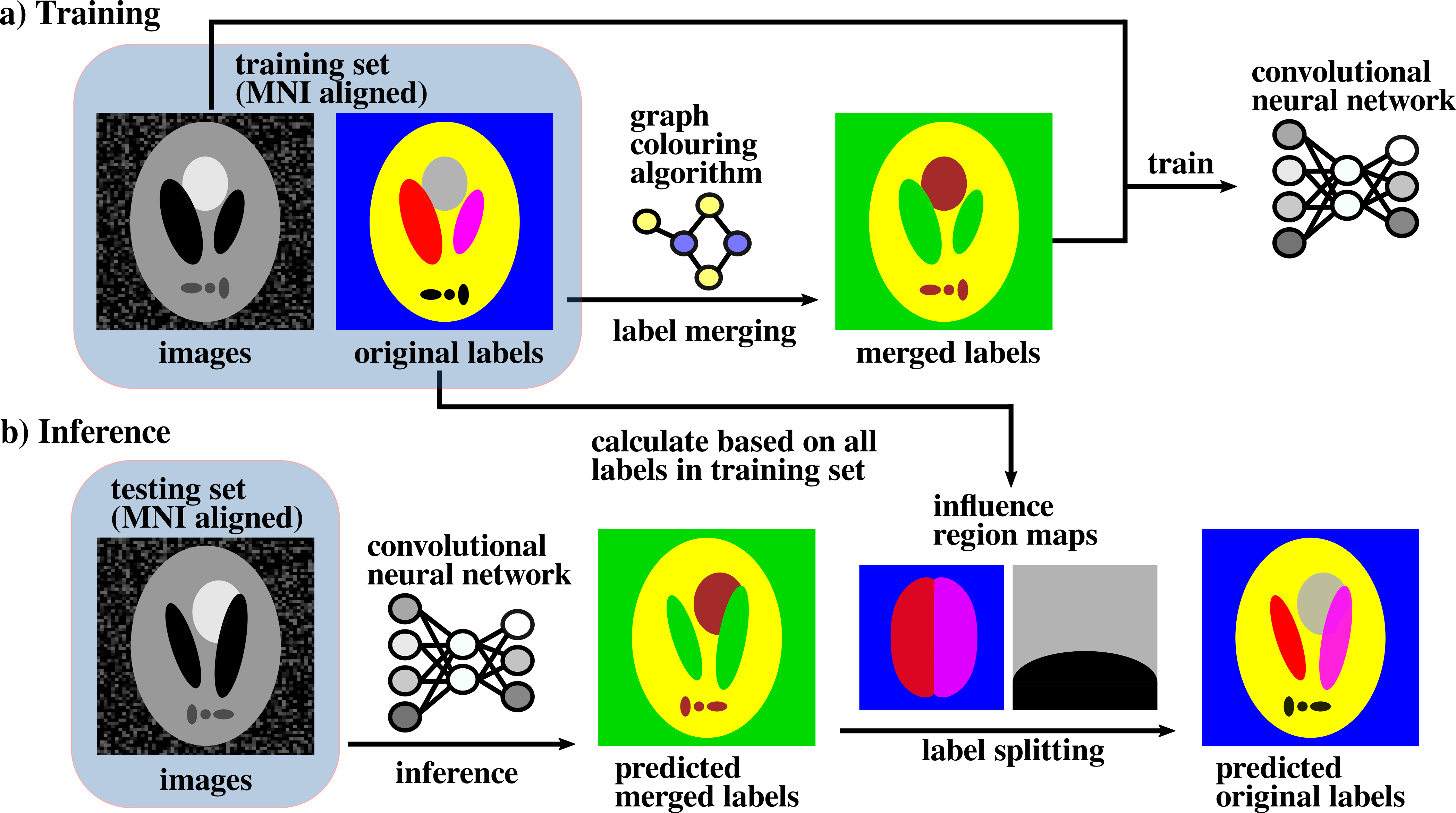}
\caption{\label{fig:splitting-merging-principle} Method overview illustrated on a toy example.
a) Training. Original labels in MNI space are automatically grouped 
using a greedy graph colouring algorithm and merged.
Toy example: \{blue, red, purple\} $\rightarrow$ green; \{gray, black\} $\rightarrow$ brown; and \{yellow\} $\rightarrow$ yellow. The CNN is trained on merged labels.
b) Inference. The CNN predicts merged labels. In a post-processing step, original labels are recovered using one influence map for every set of merged labels.}
\end{figure}

An overview of the proposed method is shown in \figref{fig:splitting-merging-principle}.

\subsubsection{Preprocessing.}
\label{sec:preprocessing}
Brain images are aligned to the MNI152 T1 template (ICBM 2009a Nonlinear Asymmetric) \cite{fonov2011unbiased} with affine registration using Advanced Normalization Tools (ANTs) \cite{avants2009advanced} and resampled to a voxel spacing of $1\times1\times1$ mm and matrix dimension $193 \times 229 \times 193$.

\subsubsection{Merging of labels.}
\label{sec:merging}
\begin{figure}[htb]
\centering
\includegraphics[width=1.0\textwidth]{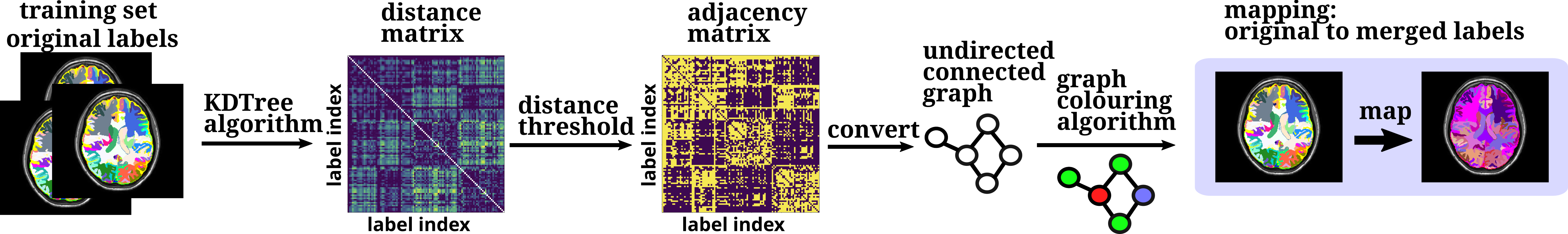}
\caption{\label{fig:merging} Label merging. The distance between original labels is calculated from (MNI aligned) training set label volumes. A distance threshold and  is applied to obtain an adjacency matrix (yellow: adjacent, blue: non-adjacent). Similarly, an average label volume ratio threshold is applied (not shown here). The adjacency matrix corresponds to an undirected connected graph, to which a greedy graph colouring algorithm is applied. Labels of the same colour are grouped, forming a new merged label.}
\end{figure}

Label merging is performed under two constraints.
\textbf{1)} The distance between to-be-merged labels in the training set must not be smaller than an empirically selected threshold to facilitate label-splitting at inference time.
\textbf{2)}~To-be-merged labels
must not be too different in volume to avoid reduced weighting of small labels in the loss function. This is achieved by setting an empirical threshold on the ratio of average label volumes in the training set. 
Our method illustrated in \figref{fig:merging} is comprised of the steps described below.

\emph{1.} To account for variability in brain morphology, we consider the support of labels in the MNI space by computing a simple prior map $S$ similar to a probabilistic atlas. 
For this, we sum the one-hot encoded training label volumes in the MNI space. Formally, let 
$X^i:\Omega \subset \mathbb{R}^3 \rightarrow \mathbb{R}$
denote the $i$th image of the set of $N_{\text{tr}}$ training images,
with $\Omega$ being the spatial domain,
and let $Y^i:\Omega \subset \mathbb{R}^3 \rightarrow \{0,\dots, N_{\text{ol}}-1\}$ be its corresponding ground-truth segmentation with $N_{\text{ol}}$ original labels.
Denoting spatial coordinates as $\mathbf{v}=(i,j,k)$ and label indices as $l \in \{0,\dots, N_{\text{ol}}-1\}$, the original label support map $S$ is given by:
\begin{align}
&S_{\mathbf{v},l} = \sum_{i=0}^{N_{\text{tr}}-1} \Tilde{Y}^i_{\mathbf{v},l}
&\text{ where } \Tilde{Y}_{\mathbf{v},l} = 
\begin{cases}
1 & \text{if } Y_\mathbf{v} = l 
\\ 
0 & \text{otherwise}
\end{cases}
\label{eq:label_support_mat}
\end{align}

\emph{2.} We calculate the minimum distances between original labels as observed in the label support map $S$.
This information is encoded in
a distance matrix $D$ 
whose entries are the minimum Euclidean distance between any two voxels of original label $l_1$ and $l_2$ taken from any two training volumes in MNI space:
\begin{equation}
D_{l_1 l_2} = \text{min}\{d(\mathbf{v}_1, \mathbf{v}_2) | \mathbf{v}_1 \in L_1, \mathbf{v}_2 \in L_2 \}
\label{eq:min_dist}
\end{equation}
where $L_1$ and $L_2$ are the sets of spatial coordinates which have non-zero support for $l_1$ and $l_2$ in the label support matrix: $L_1 = \{\mathbf{v} | S_{\mathbf{v},l_1}>0 \}$, $L_2 = \{\mathbf{v} | S_{\mathbf{v},l_1}>0 \}$.
To reduce processing time for \eqref{eq:min_dist}, it is sufficient to consider only the border voxels of $L_1$ and $L_2$.
These were extracted using the \texttt{find\_boundaries} function of the Python library scikit-image \cite{van2014scikit}. Then, a KDTree algorithm \cite{maneewongvatana1999analysis} is used on each pair of original labels and the respective boundary voxels to find the minimum distance between any two voxels of different label. Note that the matrix is symmetric, therefore calculation of the upper triangular part is sufficient.

\emph{3.}
The average volume ratio matrix $V_{l_1 l_2}$ between two labels $l_1$ and $l_2$ is calculated by dividing the larger average label volume by the smaller average label volume.

\emph{4.} We convert the distance matrix and volume ratio matrix into an adjacency matrix $A$, using the manually selected hyperparameters $\delta_{\text{D}}$ (distance threshold) and $\delta_{\text{V}}$ (volume ratio threshold):
\begin{equation}
A_{l_1 l_2} = \begin{cases} 
0 & \text{if } D_{l_1 l_2}>\delta_{\text{D}} \text{ and } V_{l_1 l_2}<\delta_{\text{V}} \\ 
1 & \text{otherwise}
\end{cases}
\label{eq:adj_mat}
\end{equation}
From the adjacency matrix $A$, an undirected graph is constructed using Python's NetworkX library \cite{hagberg2008exploring}. Each vertex of the graph represents an original label and only adjacent vertices are connected. 
Lower $\delta_{D}$ and higher $\delta_{V}$ thresholds will result in a smaller number of labels and thus greater memory savings.
Conversely, higher $\delta_{D}$ facilitates the downstream splitting of merged labels and a smaller $\delta_{V}$ prevents excessive class imbalance.

\emph{5.} Next, we cast the identification of merged label groups as a graph colouring problem.
Graph colouring assigns colours (in our case corresponding to the merged labels) to the vertices of a graph such that no two adjacent vertices are assigned the same colour.
This property, in combination with \eqref{eq:adj_mat}, ensures that only original labels with a minimum distance above $\delta_{D}$ and an average volume ratio below $\delta_{V}$ are grouped within a given merged label.
Minimising the number of merged labels, i.e. colours, will lead to minimising memory requirements. 
Finding the smallest number of colours to achieve this is an NP-hard problem \cite{garey1974some} but good practical solutions exist.
We use a greedy graph colouring method which considers the vertices of the graph in sequence and assigns each vertex its first available colour in linear time.
Specifically, NetworkX's greedy graph colouring algorithm \texttt{greedy\_color} is applied to the graph, using the “\texttt{smallest\_last}” strategy \cite{matula1983smallest}. Finally, each merged group is assigned a unique label, $\text{ml} \in \{0.. N_\text{ml}-1 \}$, where $N_\text{ml}$ is the number of merged labels.

\subsubsection{Convolutional neural network.}
\label{sec:cnn}
Our method is compatible with any neural network backbone.
Here, we rely on a patch-based 3D U-Net which acts as an established baseline.
Our implementation is based on the \texttt{DynUNet} class in the MONAI library \cite{cardoso2022monai},
itself built on top of PyTorch \cite{paszke2019pytorch}.
The network encoder and decoder paths contain 5 levels in addition to the bottleneck level.
Training policies were adapted from nnU-Net \cite{isensee2021nnu}, using random patches of size $192\times192\times128$ and a batch size of 2.
All models were trained for 1000 epochs where 1 epoch was defined as iteration over 250 mini-batches. SGD with Nesterov momentum ($\mu=0.99$) and an initial learning rate of 0.01 were used. The poly learning rate policy was employed to decrease the learning rate $(1-\frac{\text{epoch}}{\text{epoch}_\text{max}})^{0.9}$. The loss function is the sum of cross-entropy and Dice loss. In addition, deep supervision was applied in the decoder to all but the two lowest resolutions. 

\subsubsection{Splitting of labels.}
\label{sec:splitting}
\begin{figure}[tb!]
\centering
\includegraphics[width=1.0\textwidth]{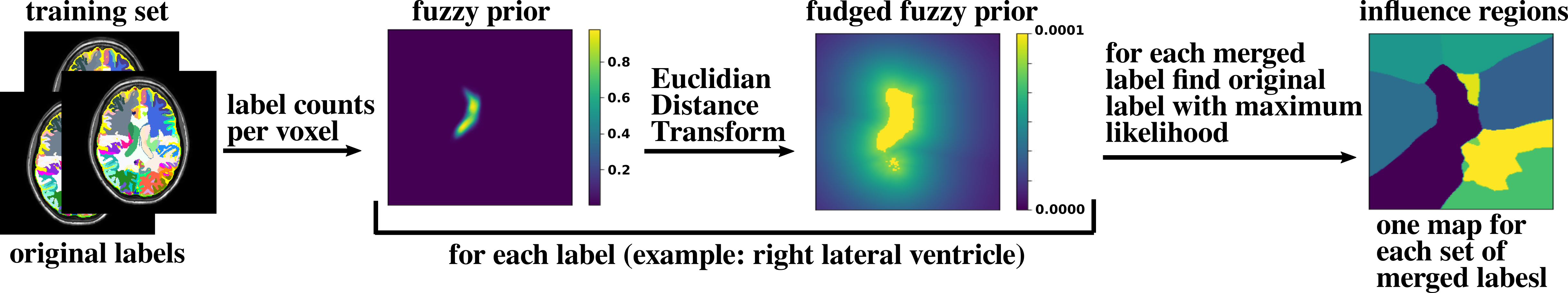}
\caption{\label{fig:splitting} Label splitting. A fuzzy prior is created from all (MNI aligned) training set label volumes. Subsequently, a Euclidean Distance Transform is applied for each label. One influence region map is created for each merged label. In this figure, the image on the right is the influence region map for the merged label that was merged from the right lateral ventricle label and 6 other original labels.}
\end{figure}

To split the merged labels predicted by the CNN, we use influence region maps created from the training set label volumes. Our splitting method illustrated in \figref{fig:splitting} is comprised of the steps described hereafter.

\emph{1.} From the original training samples, a \textit{fuzzy prior} $\hat{S}$ is created by normalizing the label support map $S$ from \eqref{eq:label_support_mat}:
$
\hat{S} = \frac{1}{N_{\text{tr}}} S
$

\emph{2.} Next, the fuzzy prior is “fudged” with an Euclidian Distance Transform (EDT) to replace entries of zero likelihood with a likelihood that reflects the spatial distance of each voxel from voxels of non-zero likelihood in $\hat{S}$. The fudged fuzzy prior is defined as:
\begin{equation}
\Tilde{S}_{\mathbf{v},l} = 
\begin{cases}
\hat{S}_{\mathbf{v},l} & \text{if }  \hat{S}_{\mathbf{v},l} > 0 \\
\frac{1}{N_{tr}} \exp(-E_{\mathbf{v},l}) & \text{otherwise} 
\end{cases}
\end{equation}
where $E_l = \text{edt}(S_l)$ is the output of the Euclidean Distance Transform. 
We use SciPy's function \texttt{distance\_transform\_edt} to calculate the EDT \cite{virtanen2020scipy}.

\emph{3.} Using the fudged prior, $N_{\text{ml}}$ influence region maps $I_{m}$ can be created. An influence region map of a merged label $m \in \{ 0..N_\text{ml}-1 \}$ is created by assigning to each voxel the original label with the highest likelihood:
\begin{equation}
I_{\mathbf{v}, m} =  \argmax_{l \in M_{m}} \Tilde{S}_{\mathbf{v},l}
\end{equation}
where $M_{m}$ is the set of original labels contained in merged label $m$.

\emph{4.} Finally, the merged labels predicted by the CNN can be split based on the influence maps. If the network predicts a merged label volume $\Tilde{N}$, each voxel's label is replaced with the original label found in the corresponding influence region map. Thus, the prediction after label splitting is given by:
\begin{align}
&\Tilde{Y} = \sum_{m=0}^{N_{\text{ml}}-1} \hat{N}{I_m}
&\text{ where } \hat{N}_{\mathbf{v},m} = 
\begin{cases}
1 & \text{if } \Tilde{N}_{\mathbf{v}} = m 
\\ 
0 & \text{otherwise}
\end{cases}
\end{align}
where $\hat{N}$ corresponds to $\Tilde{N}$ in one-hot encoded format.

\section{Evaluation methodology}
\subsubsection{Data sets.}
We evaluate the proposed method on T1-weighted images of three publicly available datasets.
\textbf{1)} Mindboggle101 dataset which contains 101 manually labelled images from various publicly available sub-datasets \cite{klein2012101}. Two of these sub-datasets (NKI-RS-22 and NKI-TRT-20) were combined into a test set of 42 images (referred to as Mindboggle42).
The remaining sub-datasets with 59 annotated images (referred to as Mindboggle59) were used as a multi-atlas for the classical GIF algorithm \cite{cardoso2015geodesic} to create pseudo-ground-truth segmentations for the AOMIC and IXI datasets.
\textbf{2)} AOMIC PIOP2 dataset which contains 226 images from healthy participants (age: 18-25 years) \cite{snoek2021amsterdam} with GIF pseudo-ground-truth labels.
\textbf{3)} IXI dataset which contains 581 images from healthy participants (age: 20-86 years) \cite{ixi} with GIF pseudo-ground-truth labels.

\subsubsection{Model training and testing.}
The AOMIC and IXI datasets were randomly split into three sets for model training ($N=144$, $N=371$), validation ($N=36$, $N=93$), and testing ($N=46$, $N=117$). Mindboggle42 was used for testing only. Mindboggle59 was not directly used for training so as to keep Mindboggle42 as a more independent test set.
To perform two independent merging and training assessments, we
used the AOMIC and IXI training datasets separately.
Following
the steps in \secref{sec:methods},
we obtained models
referred to as $\text{AOMIC}_{\text{merged}}$ and $\text{IXI}_{\text{merged}}$.
Both models are evaluated on the AOMIC, IXI, and Mindboggle42 testing sets. Preprocessing, label merging, and label splitting were performed on an Intel Core i9 CPU (8 cores at 2.3GHz). CNN training and inference were performed on 2 NVIDIA A100 GPUs with 40GB VRAM each.

\subsubsection{Baseline and evaluation metric.}
As comparison, two baseline models were trained with the original labels, i.e. without label merging and splitting. These models are referred to as $\text{AOMIC}_{\text{orig}}$ and $\text{IXI}_{\text{orig}}$. The training images and the other stages of the training and inference pipeline were identical. For evaluation, the Dice Similarity Coefficient (DSC) is applied for each label \cite{dice1945measures}. Relative volume error and Hausdorff distance are reported in the supplementary material. 

\section{Results}
\begin{figure}[htb]
\centering
\includegraphics[width=1.0\textwidth]{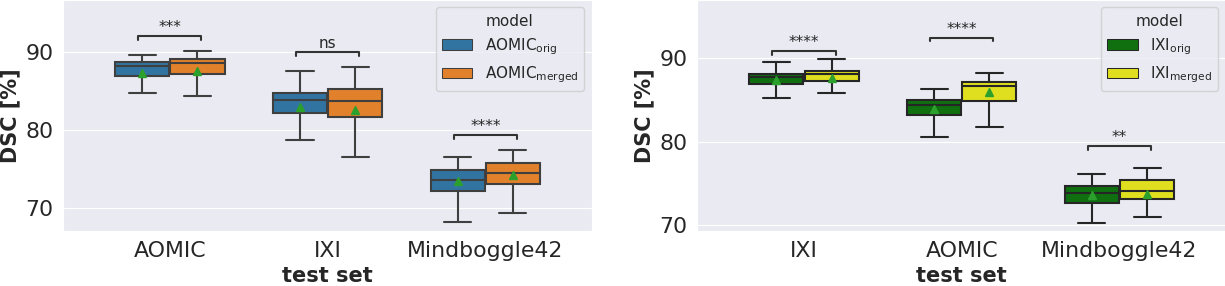}
\caption{\label{fig:boxplot_comparison_by_testset} Effect of the label merge-and-split approach on the quality of predictions. Dice Similarity Coefficients (DSC) are calculated with respect to the GIF pseudo-groundtruth (AOMIC and IXI testing sets) or manual groundtruth (Mindboggle42). Each data point is the average DSC of all labels. Boxplots with DSCs of all individual labels are included in the supplementary material. Green triangles represent the mean. Asterisks indicate statistical significance according to a Wilcoxon test (paired samples) with Bonferroni correction where *, **, ***, and **** indicate p-values below 0.05, 0.01, 0.001, 0.0001, respectively, and ns (not significant) p-values above 0.05.}
\end{figure}

The effect of hyperparameters $\delta_{\text{D}}$ and $\delta_{\text{V}}$ on the effective number of labels determined by the graph coloring algorithm is shown in Fig.~A of the supplementary material.
Values of $\delta_{\text{D}}=10$ and $\delta_{\text{V}}=3.5$ were chosen for all experiments to achieve a substantial reduction of labels while ensuring spatial separation between merged labels and avoiding high class imbalance.
The latter threshold was introduced 
after our initial experiments
showed that merging labels with a large average volume ratio resulted in reduced segmentation accuracy of small structures (supplementary material Fig.~B).
%
\tabref{tab:resources} compares the resources needed for model training and inference with and without label merge-and-split. The number of labels was reduced by a factor of at least three for both training sets. GPU memory required during training was halved and reduced by 21\% during inference. Epoch times during training were reduced by more than 43\% and inference time was halved. 
For label merging prior to model training additional pre-processing times need to be considered: calculation of the fuzzy label prior ($<$1s per training image), average volumes ($\approx$1s per training image), adjacency matrix (12min in total), and label merging before training ($\approx$2s per image) leads to a pre-processing time of less than 1h for both training sets.
At inference, label splitting takes less than 1s per image (without focus on code optimization).
Note that these additional processing times are small compared to the time required for network training/inference.
\begin{table}[htb]
\caption{GPU memory requirements and processing times for training and inference and mean Dice Similarity Coefficients (DSC) according to \figref{fig:boxplot_comparison_by_testset}}
\begin{tabular}{lrrrrrrrr}

\toprule

&\multicolumn{1}{l}{num.}&\multicolumn{1}{l}{mem.}&\multicolumn{1}{l}{mem.}&\multicolumn{1}{l}{epoch}&\multicolumn{1}{l}{infer}&\multicolumn{3}{c}{mean DSC [\%]} \\
model&\multicolumn{1}{l}{of}&\multicolumn{1}{l}{train}&\multicolumn{1}{l}{infer}&\multicolumn{1}{l}{time}&\multicolumn{1}{l}{time}& \multicolumn{3}{c}{test set} \\ \cline{7-9} 
&\multicolumn{1}{l}{labels}&\multicolumn{1}{l}{{[}GiB{]}}&\multicolumn{1}{l}{{[}GiB{]}}&\multicolumn{1}{l}{train {[}s{]}}&\multicolumn{1}{l}{{[}s{]}}&AOMIC&IXI&MB42 \\

\midrule

$\text{AOMIC}_{\text{orig}}$ & 108 & 32.1 & 3.53 & 345 & 140 & 87.2 & 82.7 & 73.1\\
$\text{AOMIC}_{\text{merged}}$ & 34 & 15.7 & 2.77 & 185 & 70 & 87.4 & 82.4 & 74.0\\
$\text{IXI}_{\text{orig}}$ & 108 & 31.9 & 3.53 & 308 & 101 & 83.8 & 87.2 & 73.4 \\
$\text{IXI}_{\text{merged}}$ & 36 & 16.1 & 2.79 & 175 & 51 & 85.8 & 87.5 & 73.5\\

\bottomrule

\end{tabular}
\label{tab:resources}
\end{table}
The effect of the merge-and-split approach on the accuracy of predictions is shown in \figref{fig:boxplot_comparison_by_testset}. 
The average DSC of the label merge-and-split models is comparable to the baseline models. In 5 of 6 comparisons the merge-and-split models even achieve statistically higher DSCs than the corresponding baseline models.

\section{Discussion and conclusion}
\label{sec:discussion}
This study shows that label merge-and-split maintains WBP segmentation performance while greatly reducing memory requirements and processing times. In contrast to previous label merging strategies that focus on merging symmetric label pairs we show that larger numbers of semantically unrelated labels can be merged and split without reducing segmentation quality. While the results presented here were obtained with the popular U-Net architecture, the method can in principle be combined with any segmentation approach to reduce the effective number of labels. 

A limitation of the proposed approach is the reliance on affine registration to a common space. Future work could focus on adapting label merge-and-split to work without affine registration. For example, the distance matrix in \eqref{eq:min_dist} for label merging can be constructed from individual distance matrices for each training case. Label splitting could be achieved by assigning original labels to connected components based on their volume and orientation with respect to each other. 

Furthermore, for applications other than WBP, the hyperparameters $\delta_{\text{D}}$ and $\delta_{\text{V}}$ might have to be optimized depending on the required number of labels. Additional constraints can be introduced in the adjacency matrix/graph to control which labels can be merged. For example, for combined WBP and lesion segmentation an additional lesion label may be introduced whose graph node is connected to all other nodes and therefore not merged with any other label.

\iftrue
\subsubsection{\discintname}
This work is supported by Medtronic Cranial \& Spinal Technologies and by core funding from the Wellcome/EPSRC [WT203148/ Z/16/Z; NS/A000049/1].
TV is supported by a Medtronic / RAEng Research Chair [RCSRF1819\textbackslash7\textbackslash34].
TV is co-founder and shareholder of Hypervision Surgical.
\fi
%
%

\bibliographystyle{splncs04}
\bibliography{Paper-2217}

\ifarXiv
    \foreach \x in {1,...,\numbersupplementpages}
    {
            \clearpage
            \includepdf[pages={\x}]{\supplementfilename}
    }
\fi

\end{document}